\documentclass[conference]{IEEEtran}
\usepackage{spconf,amsmath,epsfig}
\usepackage{multirow}
\usepackage{graphicx}
\usepackage{amsmath}
\usepackage{amssymb}
\usepackage{amsfonts}
\usepackage{booktabs}
\usepackage{textcomp}
\usepackage{graphicx}
\usepackage{algorithmic}
\usepackage{graphicx}
\usepackage{textcomp}
\usepackage{xcolor}
\usepackage{cite}
\let\OLDthebibliography\thebibliography
\renewcommand\thebibliography[1]{
  \OLDthebibliography{#1}
  \setlength{\parskip}{0pt}
  \setlength{\itemsep}{0pt plus 0.3ex}
}

\begin{document}




\title{Cycle Contrastive Adversarial Learning for Unsupervised image Deraining\\
}
%
\name{Chen Zhao, Weiling Cai$^{\dag}$, ChengWei Hu, Zheng Yuan}
\address{School of Artificial Intelligence, Nanjing Normal University, Nanjing, 210023,  China}

\maketitle

\begin{abstract}
  To tackle the difficulties in fitting paired real-world data for single image deraining (SID), recent unsupervised methods have achieved notable success. However, these methods often struggle to generate high-quality, rain-free images due to a lack of attention to semantic representation and image content, resulting in ineffective separation of content from the rain layer. In this paper, we propose a novel cycle contrastive generative adversarial network for unsupervised SID, called CCLGAN. This framework combines cycle contrastive learning (CCL) and location contrastive learning (LCL). CCL improves image reconstruction and rain-layer removal by bringing similar features closer and pushing dissimilar features apart in both semantic and discriminative spaces. At the same time, LCL preserves content information by constraining mutual information at the same location across different exemplars. CCLGAN shows superior performance, as extensive experiments demonstrate the benefits of CCLGAN and the effectiveness of its components.
\end{abstract}
\section{Introduction}
\label{sec:intro}
\small

Vision tasks, progressing notably, nonetheless confront degradation when cameras operate amidst precipitation, significantly impacting performance \cite{jiang2020multi}. Addressing this, image deraining, specifically single image deraining (SID), becomes crucial for restoring clarity. Given rain's unpredictable patterns, SID presents formidable challenges.

Initially, traditional algorithms, grounded in Gaussian mixtures \cite{li2016rain}, sparse coding \cite{luo2015removing}, and low-rank models \cite{chang2017transformed}, offered solutions. Yet, these methods, confined by rigid assumptions, struggle with rain's complexity.

Deep learning, subsequently, revolutionized deraining through extensive synthetic data training \cite{yang2017deep,fu2021rain}. Despite achievements, discrepancies between synthetic and authentic rain scenes hinder ideal outcomes.

Lack of large-scale real-world paired data exacerbates issues. Semi-supervised techniques \cite{wei2019semi,ye2021closing,yasarla2020syn2real}, blending synthetic pairs for robust initialization and unpaired real images for broader applicability, emerge. Still, domain gaps compromise effectiveness.

Our exploration targets unsupervised SID, influenced by CycleGAN innovations \cite{zhu2019singe, jin2019unsupervised, wei2021deraincyclegan}. Cycle consistency loss, however, entails "channel pollution" \cite{DBLP:conf/mm/0005000S18}, complicating interactions among image components. Additionally, maintaining content integrity and spatial accuracy during regeneration proves elusive. Thus, our first motivation is to break the cycle consistency loss. Recently, contrastive learning-based techniques have been introduced for practical deraining tasks \cite{ye2022unsupervised, chen2022unpaired}, yielding satisfactory outcomes in rain removal. Nonetheless, prior approaches overlook the critical aspect of semantic representation, leading to incomplete disentanglement of semantic details from rain layers. This oversight compromises the quality of resultant images, preventing the generation of high-caliber, rain-free visuals. Thus, our secondary incentive is to establish a semantic representation constraint, aiming to refine the separation process and enhance output fidelity.

Contrastive learning recently enhanced deraining \cite{ye2022unsupervised,chen2022unpaired}. While achieving adequacy, neglect of semantic nuances impedes thorough rain separation, compromising output quality.

Motivated, we introduce CCLGAN: a novel adversarial framework integrating cycle contrastive learning (CCL) and location contrastive learning (LCL). Our CCL, comprising intra-CCL and inter-CCL, innovates by: Intra-CCL: Crafting a semantic space conducive to superior reconstruction, while isolating rain effects by aligning rain-free images with their reconstructions, distancing these from rain-affected counterparts.
Inter-CCL: Implementing rain removal within a discriminative domain by consolidating analogous embeddings (derived rain-free information across images) and segregating dissonant ones (rain-free vs. rain-infused data).
Consequently, CCL ensures high-fidelity regeneration while adeptly excising rain across semantic and discriminative dimensions. Parallelly, LCL emphasizes content affinity, bolstering locational coherence by optimizing mutual information equivalence at corresponding input-output positions.

In summary, the main contributions of our method are as follows: 

\begin{itemize}
	\item  We introduce CCLGAN, a novel generative adversarial framework. CCLGAN proficiently eradicates rain layers, preserving critical content details. This feat is accomplished through dual strategies: cycle contrastive learning and location contrastive learning.
	
	\item Innovatively, our cycle contrastive learning encompasses twin cooperative components. Firstly, intra-CCL facilitates high-fidelity image restoration, simultaneously eliminating rain in the semantic latent domain. Secondly, inter-CCL operates within a discriminative latent space, effectively stripping away rain layers.

	\item  Comprehensive evaluations showcase CCLGAN's dominance over state-of-the-art unsupervised deraining techniques. Notably, our method competes favorably with supervised or semi-supervised alternatives. Further, extensive ablation studies affirm the efficacy of each component, underscoring their individual contributions to overall performance.
	
\end{itemize}

\begin{figure*}[t]
	\centering
	\includegraphics[width=0.9\linewidth]{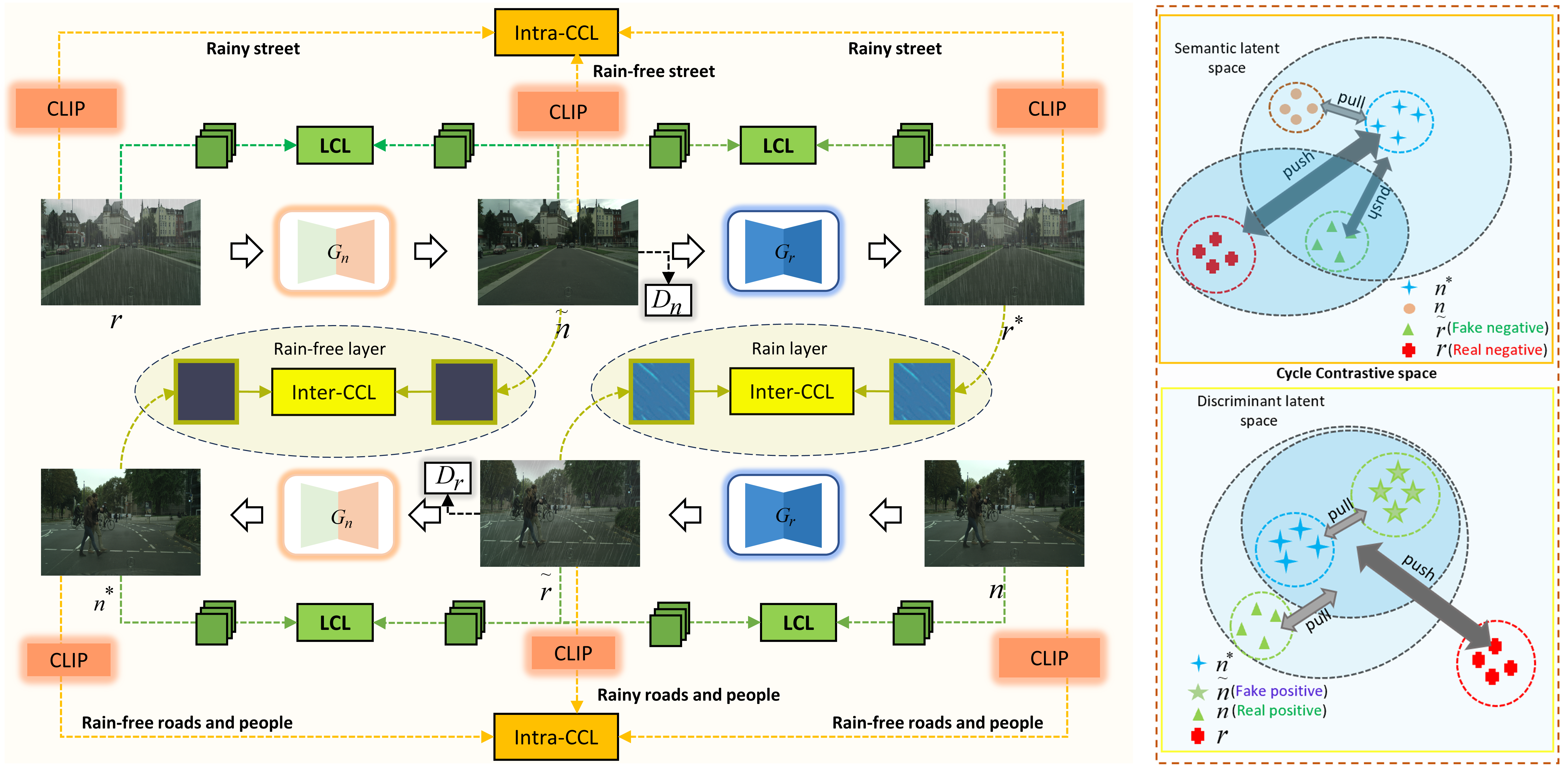}
	
	\label{fig:network}
	\caption{Overall framework of CCLGAN. It mainly consists of  cycle contrastive learning (CCL) and location contrastive learning (LCL).  
		CCL contains two cooperative branches: one is the intra-CCL  Branch and the other is the inter-CCL  Branch. Intra-CCL  Branch aims to construct a semantic latent space using Contrastive Language-Image Pre-Training (CLIP). 
		In the semantic latent space we constructed (the orange box), we pull the reconstructed rain-free image  $n^\text{*}$ and its corresponding real rain-free image $n$ close while pushing $n^\text{*}$ away from the rainy images( $\widetilde{r}$ and $r$), where the generated rainy image  $\widetilde{r}$ is fake negetive, and  the real rainy image $r$ is real negative. The fake negative $\widetilde{r}$ is similar to the query $n^\text{*}$,  making it easy to notice the rain layer for our network in the semantic latent space. 
		The real negative $r$  can make our network learn the representation of the real rain layer. Similarly, inter-CCL is proposed to realize stripping of the rain layer in a discriminative latent space. 
		In the discriminant latent space we constructed (the yellow box), pulling the reconstructed rain-free image  $n^\text{*}$ ,  the generated rain-free image $\widetilde{n}$ and the real rain-free image $n$ close and pushing them away from the real rainy images $r$ in discriminant latent space, where $n$ is real positive and $\widetilde{n}$ is fake positive. Pulling $n^\text{*}$ and $\widetilde{n}$ (fake positive) closer  aims to converge the similar feature distributions. 
		LCL aims to improve the similarity of content embeddings and enhance the location information.
		}
	\vspace{-9pt}
	
\end{figure*}
\section{Related Works}

\subsection{Single Image Deraining }
Deep learning-driven SID techniques fall into three categories: fully-supervised, semi-supervised, and unsupervised. Fully-supervised methodologies \cite{li2018recurrent,fu2021rain} demand abundant matched pairs of rainy and clean images. Pioneering work by Fu et al. \cite{fu2017removing} introduced an end-to-end residual CNN for rain streak elimination. Enhancing representation learning, subsequent strategies incorporated multi-stage \cite{yang2017deep}, multi-scale \cite{jiang2020multi}, and attention mechanisms \cite{li2018recurrent}. However, these solutions exhibit constrained versatility in real-world settings, due to a pronounced domain disparity between synthetic and genuine rainy images. Addressing this shortfall, semi-supervised deraining approaches \cite{wei2019semi,ye2021closing} emerged. Wei et al. \cite{wei2019semi}, for example, devised a semi-supervised learning paradigm for practical rain removal, scrutinizing the discrepancy between domains. More recently, unsupervised methodologies \cite{zhu2019singe,jin2019unsupervised,wei2021deraincyclegan,chen2022unpaired} harnessed enhanced CycleGAN and contrastive learning frameworks to harmonize rainy and rain-free image domains. Past efforts, however, fell short in delivering high-quality rain-free outputs, owing to insufficient emphasis on content and semantic aspects. Motivated by these observations, we aspire to engineer a potent unsupervised SID framework.

\subsection{Contrastive Learning}
Contrastive learning \cite{he2020momentum} showcases robust image representation sans labels, advancing unsupervised learning significantly. It leverages positive-negative pair dynamics, drawing queries near positives while distancing negatives in deep feature spaces. Recently, this loss function has been adapted to multiple low-level vision tasks \cite{liu2021divco,wu2021contrastive,zhao2023multi,zhao2024spectral,zhao2024learning,zhang2024novel,yuan2023context,DBLP:journals/tip/WeiLSXZ24,DBLP:conf/mm/Wei0SCX023,lou2023refining,xu2023radiology,zhang2024learnable,zheng2023cvt,zhao2024wavelet}, achieving state-of-the-art results. In this work, we innovate location contrastive learning (LCL) and cycle contrastive learning (CCL). Precisely, LCL crafts content embeddings to govern mutual information across equivalent locations in distinct samples. Meanwhile, CCL targets high-quality reconstruction and rain layer removal by mastering semantic and discriminative embeddings.

\section{Methodology}
\subsection{Overall Framework}
Our purpose is to construct a deep network for single image deraining by exploring the features from these unpaired  rainy images and clear images without the supervision of the ground truth labels. 
Due to bidirectional circulatory architecture can generate abundant rain and rain-free exemplars for unsupervised contrastive learning,  we use two generators $G_{n}$, $G_{r}$, as well as two discriminators $D_{n}$ , $D_{r}$ . The framework of CCLGAN is shown in Fig. 1, including two branches: i) rain to rain branch$\quad r\rightarrow\widetilde{n}\rightarrow r^\text{*}$, where rainy images ($ r$) are utilized to generate rain-free images ($\widetilde{n}$) and then reconstructed rainy images ($r^\text{*}$); and ii) rain-free to rain-free branch$\quad n\rightarrow\widetilde{r}\rightarrow n^\text{*}$, where rain-free images ($ n$) are employed to generate rainy images ($\widetilde{r}$) and then reconstructed rain-free images ($n^\text{*}$). 
CCLGAN mainly consists of  cycle contrastive learning (CCL) and location contrastive learning (LCL). Specifically, CCL realizes high-quality image reconstruction and stripping of the rain layer in semantic and discriminant deep spaces. Meanwhile, LCL implicitly constrains the mutual information  to maintain the content information.

\subsection{Cycle Contrastive Learning}
The proposed CCL consists of two cooperative branches: intra cycle contrastive learning (intra-CCL) Branch and inter cycle contrastive learning (inter-CCL) Branch. For intra-CCL Branch, our goal is to find a semantic latent space to achieve high-quality image reconstruction and remove the rain layer from the semantic information. Inter-CCL Branch is proposed to realize stripping of the rain layer in a discriminant latent space.

\noindent\textbf{Intra cycle contrastive learning:}
Our primary motivation is to extract semantic embeddings from images using a pre-trained classification network as the feature encoder. Then, in the semantic space, we encourage pulling the reconstructed rain-free image and the rain-free image close to each other while pushing them away from the rainy images, which constrains the generator to complete an image reconstruction and removes the rain layer from the rainy image. However, the pre-trained classification network mainly extracts the main category information of the images, which is rough semantic information. Contrastive Language-Image Pre-training (CLIP) \cite{DBLP
	/icml/RadfordKHRGASAM21}, a powerful multimodal large vision model, benefits many zero-shot learning vision tasks, which has the property of unlimited categories \cite{abs-2304-05653}.

Inspired by the work about the CLIP’s explainability \cite{abs-2304-05653}, Our aspiration is to translate images into a densely informative semantic domain. Accordingly, we fabricate a semantic latent expanse utilizing CLIP, aiming for superior image restoration whilst purging rain effects from the semantic data. Explicitly, our strategy coaxes reconstructed rain-free visuals—featuring clear pavements and pedestrians—to align closely with their authentic rain-free counterparts, simultaneously distancing themselves from rain-marred scenes, all within the semantic latent continuum. To grasp the essence of genuine rain overlays, we incorporate real-world rainy imagery as negative exemplars. The formulation for intra-cycle contrastive learning (intra-CCL) can be articulated as:

\begin{equation}
	\mathcal{L}_{intra( i)}=\dfrac{\left\|C(\boldsymbol{r}^*)-C(\boldsymbol{r})\right\|_\text{1}}{\left\|C(\boldsymbol{r}^*)-C(\widetilde{\boldsymbol{n}})\right\|_\text{1}+\left\|C(\boldsymbol{r}^*)-C(\boldsymbol{n})\right\|_\text{1}},
\end{equation}

\begin{equation}
	\mathcal{L}_{intra( ii)}=\dfrac{\left\|C(\boldsymbol{n}^*)-C(\boldsymbol{n})\right\|_\text{1}}{\left\|C(\boldsymbol{n}^*)-C(\widetilde{\boldsymbol{r}})\right\|_\text{1}+\left\|C(\boldsymbol{n}^*)-C(\boldsymbol{r})\right\|_\text{1}},
\end{equation}

\begin{equation}
	\mathcal{L}_{intra}=\mathcal{L}_{intra(i)}+\mathcal{L}_{intra( ii)},
\end{equation}
where $ i$ refers to branch i: rain to rain branch$\quad r\rightarrow\widetilde{n}\rightarrow r^\text{*}$,and $ ii$ refers to branch ii: rain to rain branch$\quad n\rightarrow\widetilde{r}\rightarrow n^\text{*}$. $r^\text{*}$ refers to 
$G_{r}(G_{n}(r))$, $n^\text{*}$ refers to  $G_{n}(G_{r}(n))$, $\widetilde{r}$ refers to $G_{r}(n)$, and $\widetilde{n}$ refers to $G_{n}(r)$. $C$ denotes that we utilize the pre-trained CLIP to extract the semantic embeddings from the images.

\noindent\textbf{Inter cycle contrastive learning:}
To guide in enhancing deraining effectiveness, inter-CCL focuses on stripping the rain layer within a discriminant space. This is achieved by promoting convergence among similar embeddings (rain-free information from different images) while ensuring that dissimilar embeddings (combining rain-free and rainy information) are pushed farther apart. However, the process of extracting discriminant embeddings through training an attention module leads to a substantial increase in training costs.As a result, we harness the discriminator's encoder directly to distill discriminative embeddings from the images. These embeddings are then forwarded to a dual-layer Multi-Layer Perceptron (MLP), tasked with projecting them into a discriminant latent space. It's noteworthy that encoder  $D_{n}$ handles rain-free images. The inter-CCL mechanism not only steers the generator towards eliminating rain-related information but also amplifies the discriminative acumen of the discriminators. The inter-CCL can be articulated as follows:

\begin{equation}
	\resizebox{0.95\linewidth}{!}{
		$\mathcal{L}_\text{n}=\dfrac{\left\|g_n(n^*)-g_n(n)\right\|_1+\left\|g_n(\widetilde{n})-g_n(n)\right\|_1+\left\|g_n(n^*)-g_n(\widetilde{n})\right\|_1}{\left\|g_n(n^*)-g_r(r)\right\|_1+\left\|g_n(\widetilde{n})-g_r(r)\right\|_1}$
	}
\end{equation}

\begin{equation}
	\resizebox{0.95\linewidth}{!}{
		$\mathcal{L}_\text{r}=\dfrac{\left\|g_r(r^*)-g_r(r)\right\|_1+\left\|g_r(\widetilde{r})-g_r(r)\right\|_1+\left\|g_r(r^*)-g_r(\widetilde{r})\right\|_1}{\left\|g_r(r^*)-g_n(n)\right\|_1+\left\|g_r(\widetilde{r})-g_n(n)\right\|_1}$
	}
\end{equation}
\begin{equation}
	\mathcal{L}_{inter}=\mathcal{L}_\text{n}+\mathcal{L}_\text{r},
\end{equation}
where $ \mathcal{L}_\text{n}$ aims to constrain $G_{n}$ to generate the rain-free information in the discriminant latent space, and  $ \mathcal{L}_\text{r}$ aims to constrain $G_{r}$ to generate the real rainy information. $g_n$ and $g_r$  denotes that we utilize the $D_{n}$ and $D_{r}$ as encoder to extract the rain-free and rainy embeddings from the images, respectively.

\subsection{Location Contrastive Learning}

LCL aims to maintain content information from rainy images and improve location details during the image deraining process. We observe that corresponding locations in the rainy image $\quad r$ and the generated rain-free image $\widetilde{n}$ are visually similar to each other. Thus our purpose is  to maximize mutual information between input and output of generator in the same location. We extract content feature from the input image by encoder of the generator. Then, we map the content feature into the content deep space via a  two-layer MLP.Note that, we use encoder 
$G_{n}$ for the rainy image, and we use encoder  $G_{r}$ for the rain-free image. Additionally, separate MLPs are employed for mapping the content features of the rainy and rain-free images. Patches from the same location between input and output are treated as positives, while patches from different locations serve as negatives in our formulation of location contrastive loss.

\begin{equation}
	\resizebox{0.95\linewidth}{!}{
		$\mathcal{L}_{LCL}=-\log\left[\dfrac{\exp\left(Sim(q, k^{+})\ /\tau\right)}{\exp\left(Sim(q, k^{+})\
			/\tau\right)+\sum_{i=1}^N\exp\left(Sim(q, k^{-})\
			/\tau\right)}\right]$
	}
\end{equation}
where  $Sim(u^\top, v)=u^\top v/\text{}\|u\|\|v\|$ denotes the cosine similarity between $ u$ and $ v$. $ q$ refers to the a query from the output, $ k^{+}$ refers to a positive and $k^{-}$ refer to  negatives from the input. For example, the house (query) in the output should be more similar  to the house in the input image, but far away the road in the content deep space. $N$ is the negative sample numbers. 
Note that the positive $ k^{+}$ is the corresponding of the query $ q$, and N negatives $ k^{-}$ are randomly selected in the input. Here $\tau$ indicates a temperature parameter used to measure the distance between query and other samples.

We define the adversarial loss $\mathcal{L}_{adv}$ as
\begin{eqnarray}
	\mathcal{L}_{adv_n}=\mathbb{E}_{n\in N}\log D_{n}(\mathbf{n})+\mathbb{E}_{r\in R}\log(1-D_{n}(G_{n}(\mathbf{r}))),
\end{eqnarray}
\begin{eqnarray}
	\mathcal{L}_{adv_r}=\mathbb{E}_{r\in R}\log D_{r}(\mathbf{r})+\mathbb{E}_{n\in N}\log(1-D_{r}(G_{r}(\mathbf{n}))),
\end{eqnarray}
\begin{eqnarray}
	\mathcal{L}_{adv}=	\mathcal{L}_{adv_n}+	\mathcal{L}_{adv_r},
\end{eqnarray}
where $\mathcal{L}_{adv}$ constrains $G_{n}$ and $G_{r}$  to make the generated images  similar to real samples. In contrast, $D_{n}$ and $D_{r}$ are exploited to distinguish the generated images and real images.

In summary, the overall objective function can be formulated as:

\begin{eqnarray}
	\mathcal{L}_{total}=\mathcal{L}_{adv}+\lambda_{\text{1}}\mathcal{L}_{_{LCL}}+\lambda_{\text{2}}\mathcal{L}_{intra}+\lambda_{\text{3}}\mathcal{L}_{inter},
\end{eqnarray}
where $\lambda_{\text{1}}$, $\lambda_{\text{2}}$, $\lambda_{\text{3}}$ and $\lambda_{\text{4}}$ are trade-off parameters. We set $\lambda_{\text{1}}=0.5$  , $\lambda_{\text{2}}=0.5$, and $\lambda_{\text{3}}=0.05$.

\begin{figure*}[t]
	\centering
	\includegraphics[width=1\linewidth]{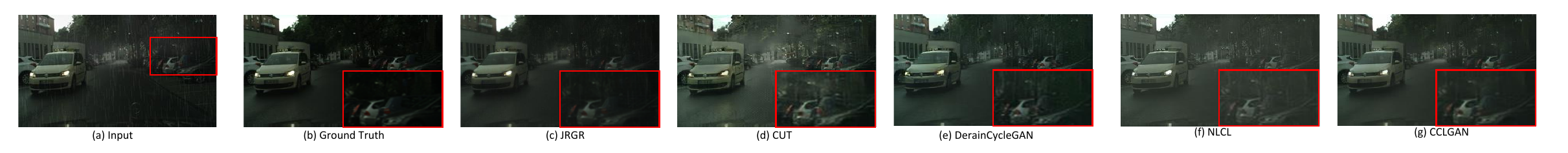}
	\caption{Visual comparison on  RainCityscapes.  In contrast, our methods achieve more natural results.} 
	\label{fig:2}
\end{figure*}

\begin{figure*}[t]
	\centering
	\includegraphics[width=1\linewidth]{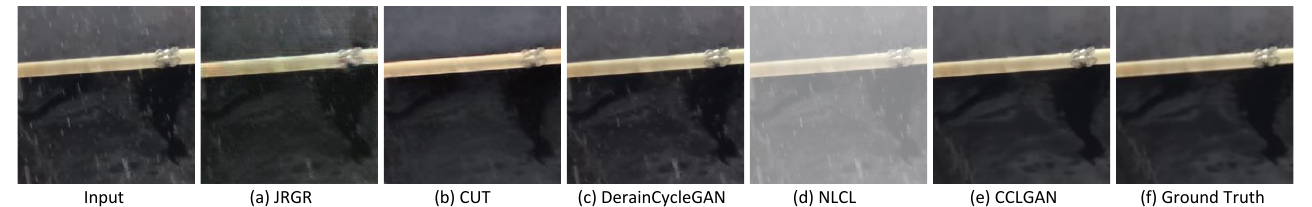}
	\caption{Visual comparison on  SPA.  In contrast, our methods achieve better  results.} 
	\label{fig:2}
\end{figure*}

\begin{table}
	\caption{Quantitative comparisons of different methods on synthetic and real datasets. Sup, Semi and Semi mean supervised, semi-supervised and unsupervised methods, respectively.}
	\label{tab:freq}
	\resizebox{0.95\linewidth}{!}{
	\begin{tabular}{cc|cc|cc}
		\toprule
		
		\multicolumn{2}{c|}{Datasets}                      & \multicolumn{2}{c|}{RainCityscapes} &  \multicolumn{2}{c}{SPA} \\
		
		\midrule
		\multicolumn{2}{c|}{Metrics}  & PSNR    & SSIM  & PSNR &SSIM          \\\midrule
		Sup& DDN & 24.12         & 0.8781                & 33.74         & 0.9112        \\
		Sup& RESCAN & 24.89         &  0.9101                & 33.11         & 0.9253        \\
		Sup& JORDER-E  &25.64 &0.8767& 33.28& 0.9406  \\
		\midrule
		Semi& SSIR & 25.08& 0.8853& 30.78& 0.8668  \\
		Semi& Syn2Real&  25.32 &0.8871& 33.14 &0.9183  \\
		Semi&  JRGR &27.51 &0.9132& 35.59& 0.9498 \\
		\midrule
		Unsup& CycleGAN& 24.86& 0.7906&  33.54& 0.9127   \\
		Unsup&CUT& 25.21& 0.8225 & 32.97& 0.9434    \\
		Unsup&UDGNet& 25.16& 0.8749&  29.67& 0.9299    \\
		Unsup&DeCycleGAN&  26.99& 0.8670&  34.16 &0.9436   \\
		Unsup&NLCL&  26.46& 0.8666&  33.82& 0.9468                 \\
		Unsup&DCD-GAN&  25.18& 0.8270&  29.23& 0.9195                 \\
		Unsup&ANLCL& 27.42 &0.9123&  \textbf{35.07}& 0.9505                 \\
		\midrule
		Unsup&\textbf{CCLGAN}& \textbf{29.17} &\textbf{0.9234}&  34.51& \textbf{0.9513}                 \\

		\bottomrule
	\end{tabular}
}
\end{table}


\section{Experiments}

\subsection{Implementation Details and Methods}
\noindent \textbf{Implementation details}. For fair comparison, the implementation of CCLGAN is mainly based on the CycleGAN \cite{zhu2017unpaired}, where we use a ResNet-based generator with 9 residual blocks and a PatchGAN discriminator \cite{johnson2016perceptual,isola2017image}. During training, we use the Adam optimizer, $\beta_{1}= 0.5$ and $\beta_{2}= 0.999$. Within LCL and inter-CCL, our configuration involves deriving multi-layer attributes from the generator and discriminator encoders, acquiring content and discriminative multi-layer embeddings \cite{park2020contrastive}. Employing a batch dimension of 1, all training visuals are scaled to dimensions of 286 × 286, subsequently cropped into segments measuring 256 × 256. CCLGAN undergoes 400 epochs of training per dataset, with an initial learning rate set at 0.0002. Post 200 epochs, the learning rate diminishes linearly toward zero. Execution of the entire neural network is facilitated using PyTorch 1.7, powered by an Nvidia GeForce RTX 3090 Graphics Processing Unit.

\noindent \textbf{Comparison methods}. We compare our method with three supervised methods (DDN \cite{fu2017removing}, RESCAN \cite{DBLP:conf/eccv/LiWLLZ18} and JORDER-E \cite{yang2019joint}), three semi-supervised approaches (SSIR \cite{wei2019semi}, Syn2Real \cite{yasarla2020syn2real} and JRGR \cite{ye2021closing} ), and six unsupervised deep nets ( CycleGAN \cite{zhu2017unpaired}, CUT \cite{park2020contrastive}, UDGNet \cite{DBLP:conf/mm/Yu00ZY21}, DerainCycleGAN \cite{wei2021deraincyclegan}, NLCL \cite{ye2022unsupervised}, DCD-GAN \cite{chen2022unpaired} and ANLCL \cite{anlcl}). Note that we can only learn our networks in an unsupervised manner.

\subsection{ Datasets and Evaluation Metrics}
Utilizing two demanding benchmark collections, we assess our methodology: the synthetic RainCityscapes dataset \cite{hu2019depth} and the real-world SPA dataset \cite{wang2019spatial}. Concerning RainCityscapes, our training phase exploits 1400 images, complemented by 175 images reserved for evaluation. Regarding SPA, we allocate 2000 images for training and designate 200 images for testing purposes. Primarily, we invoke widely-recognized full-reference image quality evaluation measures: Peak Signal-to-Noise Ratio (PSNR) and Structural Similarity Index (SSIM) \cite{DBLP:journals/tip/WangBSS04}. PSNR and SSIM respectively gauge pixel-level and structural-level discrepancies between our outcomes and competing methods. Higher PSNR and SSIM values unequivocally indicate superior image quality, reflecting the efficacy of our approach.

\subsection{Comparison with other Methods}

Table 1 displays quantitative outcomes against all benchmarks on RainCityscapes and SPA, encompassing three supervised methods, three semi-supervised approaches, and six unsupervised deep nets. Our primary metrics are PSNR and SSIM. The findings reveal our algorithms outperform state-of-the-art approaches and achieve top performance on SID, affirming the robustness of the proposed CCLGAN.

To further illustrate our superiority, Figs. 2 and 3 offer visual comparisons with contemporary methods on RainCityscapes and SPA. Our techniques preserve structural information and produce high-quality images with natural details, robustly confirming their efficacy. Additionally, Fig. 4 assesses performance on real rainy scenes, demonstrating our methods yield natural and visually superior outcomes by effectively eliminating rain streaks and veiling artifacts.
\begin{table}
	\caption{Effectiveness of  each  loss function in CCLGAN.}
	\label{tab:freq}
	\resizebox{0.95\linewidth}{!}{
		\begin{tabular}{c|cccc|cc}
			\toprule
			
			Method  & $\mathcal{L}_{adv}$    & $ \mathcal{L}_{LCL} $& $\mathcal{L}_{intra}$& $\mathcal{L}_{inter}$& PSNR &SSIM          \\\midrule
			
			A&$\times$   & $\checkmark$        &  $\checkmark$             & $\checkmark$         & 26.61 &  0.8789      \\
			B&$\checkmark$  & $\times$        &  $\checkmark$              & $\checkmark$         & 26.45 & 0.8771       \\

			C&$\checkmark$    & $\checkmark$        &  $\times$                & $\checkmark$        &  27.67& 0.8860       \\
			D&$\checkmark$  & $\checkmark$        &  $\checkmark$              & $\times$         & 27.43 &   0.9117     \\
			\midrule
			E&$\times$  & $\checkmark$        &  $\times$              & $\times$         & 26.37 &0.8498        \\
			
			F&$\times$  & $\times$        &  $\checkmark$              & $\checkmark$         & 26.21 &0.8544        \\

			\midrule
			G&$\checkmark$   & $\checkmark$        &  $\checkmark$             & $\checkmark$         & \textbf{29.17} & \textbf{0.9234 }      \\
			
			\bottomrule
	\end{tabular}}
\vspace{-15pt}
\end{table}

\subsection{Ablation Study}
\noindent\textbf{Effectiveness of  each loss function in CCLGAN:} To assess our contrastive loss functions, we conduct an ablation study on RainCityscapes in Table 2. Initially, we systematically remove one component from each configuration. Model G achieves optimal performance using all components, demonstrating each loss contributes uniquely to our framework. In terms of SSIM, Model D surpasses A, B, and C, illustrating the effectiveness of our intra-CCL in enhancing structural information.
Furthermore, we investigate our LCL separately, and Model E outperforms CycleGAN, indicating LCL effectively preserves content information. Model F achieves satisfactory results, affirming our CCL enhances image reconstruction and rain removal effectively.

\noindent\textbf{Strategy of selecting negatives and positives:} In this section, we explore strategies for intra-CCL and inter-CCL separately. Concerning intra-CCL, the key design factor is how negatives are chosen. Table 3 presents quantitative findings on RainCityscapes, illustrating our method achieves superior PSNR and SSIM values. Solely using real negatives results in inadequate performance because their semantic embeddings diverge from the query, complicating rain layer detection in the semantic latent space. Conversely, relying exclusively on fake negatives fails to capture the representation of the actual rain layer.
Table 4 exhibits quantitative outcomes for selecting positives in inter-CCL, where our approach achieves optimal results. We propose that aligning reconstructed and generated rain-free information promotes aggregation of similar discriminant latent space features, facilitating effective recognition by the generator and discriminator.

\section{Conclusion}
 In this paper,we unveil a pioneering cycle contrastive learning generative adversarial framework, tailored for efficient unsupervised single image deraining. Our devised algorithm demonstrates state-of-the-art performance. Through exhaustive trials, we elucidate the potency of each component; empirical evidence corroborates their effectiveness. Indeed, our conceptualized framework embodies a universal unsupervised strategy within low-level vision tasks. As a result, future explorations will target its application breadth, investigating capabilities in alternative domains such as underwater image enhancement, atmospheric haze removal, and noise mitigation. This endeavor will underscore the framework's versatility and robustness across various imaging challenges.

\begin{table}
	\caption{Strategy of selecting negatives in intra-CCL.}
	\label{tab:freq}
	\centering
	\begin{tabular}{cccc}
		\toprule
		
		Fake Negative  & Real Negative    &  PSNR &SSIM          \\\midrule
		
		$\times$   & $\checkmark$       & 27.12 &0.8891        \\
		$\checkmark$  & $\times$         & 27.33 &0.8988        \\
		$\checkmark$    & $\checkmark$       &  \textbf{27.43}& \textbf{0.9117}       \\
		\bottomrule
	\end{tabular}
\vspace{-15pt}
\end{table}

\begin{table}
	\caption{Strategy of selecting positives in inter-CCL.}
	\label{tab:freq}
	\centering
	\resizebox{0.8\linewidth}{!}{
	\begin{tabular}{cccc}
		\toprule
		
		Fake Positive  & Real Positive    &  PSNR &SSIM          \\\midrule
		
		$\times$   & $\checkmark$       & 27.55 &0.8832        \\
		$\checkmark$  & $\times$         & 26.69 &0.8799       \\
		$\checkmark$    & $\checkmark$       &  \textbf{27.67}& \textbf{0.8860}       \\
		\bottomrule
	\end{tabular}}
\vspace{-15pt}
\end{table}

{
	\scriptsize
\bibliographystyle{IEEEbib}
\bibliography{icme2022template}
}
\end{document}